\documentclass[11pt]{article}

\usepackage[preprint]{acl}

\usepackage{times}
\usepackage{latexsym}

\usepackage[T1]{fontenc}

\usepackage[utf8]{inputenc}

\usepackage{microtype}

\usepackage{inconsolata}

\usepackage{graphicx}

\usepackage{amsmath}
\usepackage{pifont}
\usepackage{booktabs}
\usepackage{multirow}
\usepackage{amssymb}
\usepackage{algorithm}
\usepackage{algorithmic}  
\usepackage{caption}
\usepackage{bm}
\usepackage{xspace}
\usepackage{pifont}
\usepackage[table,x11names]{xcolor}
\usepackage{cleveref}
\usepackage{enumitem}
\usepackage{booktabs}     
\usepackage{multirow}      
\usepackage{graphicx}     
\usepackage{array}        
\usepackage{makecell}    
\usepackage{caption}      
\usepackage{adjustbox}   
\newcommand{\sysname}{ConfSpec\xspace}

\definecolor{darkgreen}{RGB}{0,140,90}
\definecolor{darkred}{RGB}{160,0,0}

\title{\sysname: Efficient Step-Level Speculative Reasoning via Confidence-Gated Verification}

\author{
Siran Liu$^{1,2}$,  Cyril Y. He$^{2}$\thanks{Corresponding Author.} \\ 
\normalsize $^1$Peking University, $^2$ScitiX AI.\\
\normalsize \texttt{liusr25@stu.pku.edu.cn} \\
}

\begin{document}
\maketitle
\begin{abstract}
Chain-of-Thought reasoning significantly improves the performance of large language models on complex tasks, but incurs high inference latency due to long generation traces. Step-level speculative reasoning aims to mitigate this cost, yet existing approaches face a long-standing trade-off among accuracy, inference speed, and resource efficiency.

We propose \sysname, a confidence-gated cascaded verification framework that resolves this trade-off. Our key insight is an asymmetry between generation and verification: while generating a correct reasoning step requires substantial model capacity, step-level verification is a constrained discriminative task for which small draft models are well-calibrated within their competence range, enabling high-confidence draft decisions to be accepted directly while selectively escalating uncertain cases to the large target model.

Evaluation across diverse workloads shows that \sysname achieves up to 2.24$\times$ end-to-end speedups while matching target-model accuracy. Our method requires no external judge models and is orthogonal to token-level speculative decoding, enabling further multiplicative acceleration.
\end{abstract}
\section{Introduction}

Large language models (LLMs)~\cite{liu2024deepseek, achiam2023gpt} have demonstrated remarkable reasoning capabilities by explicitly generating chains of thought (CoT)~\cite{wei2022chain}, enabling strong performance on complex tasks such as mathematical problem solving, scientific question answering, and program synthesis.
By decomposing a problem into a sequence of intermediate reasoning steps, CoT allows models to perform multi-step inference that would otherwise be infeasible with direct answer generation.
However, this expressive reasoning ability comes at a substantial computational cost. Modern reasoning-oriented models often produce thousands of tokens per example, making inference latency and serving cost a critical bottleneck for large-scale deployment~\cite{pan2025specreason, fu2025scaling}.

Speculative decoding~\cite{leviathan2023fast} mitigates this overhead by allowing a lightweight draft model to propose candidate tokens that are verified by a larger target model.
While effective at the token level, this paradigm overlooks a key structural property of CoT reasoning: reasoning progresses through semantically coherent \emph{steps} rather than isolated \emph{tokens}.
Token-level verification enforces strict surface-form equivalence, causing semantically valid but lexically different continuations to be frequently rejected, and leaving substantial efficiency gains untapped.

Motivated by this observation, recent work explores \emph{step-level speculative reasoning}, where entire reasoning steps are proposed and verified at once. However, existing approaches face a fundamental limitation that goes beyond a simple speed--accuracy trade-off. Scoring-based methods~\cite{pan2025specreason} rely on lightweight heuristics that implicitly assume optimism in draft generation, often accepting fluent but logically incorrect steps. Lookahead-based approaches~\cite{fu2025scaling} rely on brute-force verification by invoking additional judge models, effectively stacking model capacity rather than exploiting structural asymmetries. Embedding-based verification~\cite{fu2025scaling} reduces overhead but collapses rich logical structure into low-dimensional similarity, leading to brittle semantic judgments. As a result, current methods exhibit a persistent tension among accuracy, inference speed, and resource efficiency, as summarized in Table~\ref{tab:tradeoff}.

\begin{table}[t]
\centering
\small
\caption{Comparison of speculative reasoning approaches under the accuracy--speed--resource trade-off.}
\begin{tabular}{lccc}
\toprule
\textbf{Method}
& \textbf{Accu.}
& \textbf{Speed}
& \textbf{Res. Effi.} \\
\midrule
\makecell[l]{Scoring-Based}
& \textcolor{darkred}{\ding{55}}
& \textcolor{darkgreen}{\ding{51}}
& \textcolor{darkgreen}{\ding{51}} \\

\makecell[l]{Lookahead-Based}
& \textcolor{darkgreen}{\ding{51}}
& \textcolor{darkgreen}{\ding{51}}
& \textcolor{darkred}{\ding{55}} \\

Embedding-Based
& \textcolor{darkred}{\ding{55}}
& \textcolor{darkgreen}{\ding{51}}
& \textcolor{darkgreen}{\ding{51}} \\
\midrule
\textbf{\sysname(Ours)}
& \textcolor{darkgreen}{\ding{51}}
& \textcolor{darkgreen}{\ding{51}}
& \textcolor{darkgreen}{\ding{51}} \\
\bottomrule
\end{tabular}
\label{tab:tradeoff}
\vspace{-1mm}
\end{table}

Across these methods, \emph{step-level verification is implicitly treated as uniformly hard}, an assumption that leads to either unnecessary computation or irreversible accuracy loss. In this work, we argue that the true bottleneck of step-level speculative reasoning lies not in step generation, but in overly conservative verification, which forces expensive model verification even for routine reasoning steps. This bottleneck stems from a fundamental asymmetry: while generating a correct reasoning step often requires substantial model capacity, verifying whether two candidate steps are semantically equivalent is typically easier~\cite{zhou2025variation} and highly uneven in difficulty~\cite{ding2024easy2hard}. Many verification instances involve routine arithmetic or local logical transformations, while only a small fraction require deep semantic scrutiny.

Crucially, and contrary to the common belief that small language models are poorly calibrated or prone to overconfidence, we make a counter-intuitive observation:
\emph{for the specific task of step-level verification, small models remain surprisingly well-calibrated within their competence range} (validated in \S\ref{sec:preliminary}).
When a draft model assigns high confidence to a verification decision, that decision is usually correct.
Low confidence, in contrast, reliably signals genuinely ambiguous or out-of-distribution cases.
This property suggests that heavyweight target-model verification is unnecessary for the majority of reasoning steps and should be reserved for difficult instances.

Based on these insights, we propose \sysname, a confidence-gated cascaded verification framework for step-level speculative reasoning.
Instead of uniformly invoking the target model for verification, \sysname introduces a two-stage cascade.
The draft model first performs semantic equivalence verification and produces both a decision and a confidence score.
High-confidence decisions are accepted directly, while low-confidence cases are escalated to the target model for definitive verification.
This confidence-aware routing enables efficient use of model capacity without sacrificing correctness.

We evaluate \sysname on mathematical reasoning, scientific QA, and code generation benchmarks, which require long chains of intermediate reasoning and represent diverse reasoning paradigms, including AIME, AMC, MATH, GPQA, and HumanEval (\S\ref{sec:experiments}). Across tasks and model families, \sysname achieves up to 2.24$\times$ inference speedup while matching target-model accuracy. Our approach requires no external judge models and is fully compatible with token-level speculative decoding, enabling multiplicative speedups. These results demonstrate that verification-level cascading provides a principled solution to the long-standing speed--accuracy--resource trade-off in step-level speculative reasoning.

\section{Background}

\subsection{Reasoning with LLMs}

LLMs perform complex reasoning through the generation of CoT.
Given an input prompt $x$, a model $M$ produces a sequence of intermediate reasoning steps
$S = \{s_1, s_2, \ldots, s_n\}$ before deriving the final answer $y$.
Each reasoning step $s_i$ consists of multiple tokens and represents a semantically coherent logical transformation.

In this work, we define a \emph{reasoning step} as a contiguous span of tokens delimited by a step boundary token. Concretely, we adopt newline-separated steps, where each step ends with newline token ``$\backslash$n$\backslash$n''. This convention aligns with the formatting of long-CoT reasoning traces~\cite{wei2022chain} in modern reasoning-oriented LLMs and enables explicit step-level generation and verification. Both draft and target models generate steps autoregressively until a step boundary is reached.

Although reasoning traces are generated token by token, their semantic structure is naturally organized at the step level.
This mismatch between the unit of generation and the unit of semantics motivates acceleration methods that operate beyond individual tokens.

\subsection{Token-Level Speculative Decoding}

Speculative decoding~\cite{leviathan2023fast} accelerates autoregressive generation through a draft-then-verify paradigm.
A lightweight draft model $M_D$ first generates $k$ candidate tokens, which are then verified by a larger target model $M_T$ in a single batched forward pass.
Token acceptance is determined via rejection sampling based on probability ratios, guaranteeing that the resulting output distribution exactly matches that of $M_T$.

While effective for general text generation, speculative decoding enforces strict token-wise equivalence. In reasoning tasks, however, multiple token sequences may express the same underlying logic. Consequently, semantically valid draft continuations are frequently rejected due to surface-form discrepancies, limiting the achievable speedup~\cite{pan2025specreason}.

\subsection{Step-Level Speculative Reasoning}

Step-level speculative reasoning extends speculative decoding by operating on entire reasoning steps.
By allowing verification based on semantic equivalence rather than exact token matching, step-level speculation better aligns with the structure of CoT reasoning.

This shift introduces a central challenge: \emph{step-level verification}.
Formally, given a draft step $\hat{s}_i$ and a target step $s_i$, verification determines whether the two steps are semantically equivalent.
We denote this decision by a verification function $V(\hat{s}_i, s_i) \in \{\texttt{accept}, \texttt{reject}\}$.

Strict equality requires $\hat{s}_i = s_i$ at the token level, which is unnecessarily restrictive.
In contrast, semantic equivalence captures the notion that two steps induce similar downstream reasoning behavior.
One possible formalization is that $\hat{s}_i$ and $s_i$ are equivalent if they induce similar conditional distributions over future outputs, i.e., $
P(y \mid x, s_1, \ldots, s_{i-1}, \hat{s}_i) \approx
P(y \mid x, s_1, \ldots, s_{i-1}, s_i)$.
Designing an efficient and reliable approximation to this equivalence test remains the key challenge in step-level speculative reasoning.

\section{Related Work}

\subsection{Token-Level Speculative Decoding}

Speculative decoding has been extensively studied as a general mechanism for accelerating autoregressive LLM inference.
Existing approaches differ in how draft tokens are generated and verified, including standalone draft models, auxiliary decoding heads (e.g., Medusa, Hydra~\cite{cai2024medusasimplellminference,ankner2024hydrasequentiallydependentdraftheads}), feature-aligned draft models (e.g., EAGLE~\cite{li2025eaglespeculativesamplingrequires,li2025eagle3scalinginferenceacceleration}), retrieval-based methods~\cite{saxena2023prompt,fu2024breaksequentialdependencyllm}, and tree-based parallel speculation~\cite{Miao_2024}.
Additional optimizations reuse the target model's KV cache to reduce redundant computation~\cite{du2024glidecapelowhasslemethod,zimmer2025mixtureattentionsspeculativedecoding}.

However, strict token-wise or distribution-level matching causes semantically equivalent reasoning trajectories to be rejected due to superficial differences, motivating higher-level speculation units that tolerate semantic variation.

\subsection{Step-Level Speculative Reasoning}

To address token-level rigidity, recent work explores speculative reasoning at the granularity of reasoning steps.
SpecReason~\cite{pan2025specreason} employs a lightweight speculator to propose an entire step, which is scored by the target model in a single prefill pass.
While computationally efficient, this approach often sacrifices accuracy by accepting high-quality but semantically incorrect steps.

LookaheadReasoning~\cite{fu2025scaling} improves verification fidelity by generating multiple future steps and invoking an external LLM-as-a-Judge~\cite{zheng2023judging} to assess semantic equivalence.
However, this design incurs substantial resource overhead—requiring deployment of an additional inference engine for the judge model—and introduces redundancy: the model already possesses strong verification capability through its internal representations and output probabilities.
Embedding-based alternatives~\cite{fu2025scaling} employ lightweight embedding models for step alignment to reduce cost, yet compress rich logical structure into low-dimensional similarity metrics, leading to unreliable alignment for complex reasoning.

These approaches highlight a persistent trade-off among accuracy, speed, and resource efficiency, suggesting that the core challenge lies in verification rather than speculation granularity.

\subsection{Uncertainty and Self-Verification in LLMs}

\sysname is closely related to work on uncertainty estimation, calibration, and self-verification in LLMs. Prior studies~\cite{jiang2020can, kadavath2022language} show that model output probabilities encode meaningful uncertainty signals and that LLMs can often identify when they are uncertain or likely to be incorrect. Such confidence estimates have been used for selective prediction, abstention, self-consistency, and iterative self-correction.

Unlike prior work that relies on external judges or additional training objectives, \sysname leverages intrinsic confidence estimates from the draft model to guide verification. This design aligns with findings~\cite{zhou2025variation, jiang2020can, saad2025shrinking} that, for discriminative tasks within a model’s competence range, confidence scores can serve as reliable indicators of correctness. By exploiting this property, \sysname achieves intrinsic verification without introducing additional models or supervision.

\begin{figure}[ht]
  \centering
  \includegraphics[width=\linewidth]{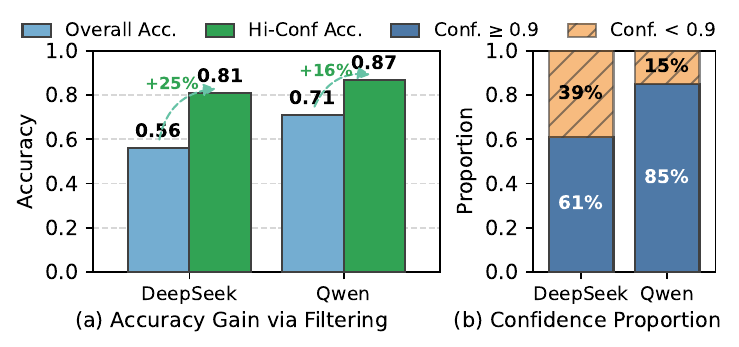}
 \caption{Confidence-based filtering improves verification accuracy. (a) Restricting to high-confidence predictions ($p_D \ge 0.9$) substantially improves draft-model accuracy, approaching target-model quality. (b) High-confidence cases cover the majority of verification instances (61--85\%), enabling efficient cascading.}
  \label{fig:motivation}
  \vspace{-1mm}
\end{figure}

\section{Methodology}
\label{sec:method}

We introduce \sysname, a confidence-gated cascaded verification framework for step-level speculative reasoning.
The core idea is to exploit an empirical asymmetry in verification difficulty: most reasoning steps can be verified reliably by lightweight draft model, while only a small fraction require expensive target-model verification.
We first present a concise preliminary study motivating confidence-based routing, then formally define the \sysname framework, followed by an extension to tree-structured drafting and an efficiency analysis.

\subsection{Confidence-Guided Verification}
\label{sec:preliminary}

Our design is motivated by a critical asymmetry in step-level verification, which gives rise to a key empirical finding: model confidence reliably signals verification difficulty.

\noindent
\textbf{Verification Asymmetry and Heterogeneity.}
In practice, step-level verification exhibits a strong asymmetry relative to generation. Judging whether two reasoning steps are semantically equivalent reduces to a constrained discriminative decision, whereas generating a valid step requires exploring a substantially larger output space. Moreover, verification difficulty is highly heterogeneous: many instances involve routine arithmetic or local logical transformations, while only a small fraction require deeper semantic reasoning. Together, these properties suggest that a lightweight model can handle a large fraction of verification cases reliably.

\noindent
\textbf{Confidence as a Reliability Signal.}
Crucially, we observe that for step-level verification, the draft model's confidence strongly correlates with correctness. 
Let $p_D$ denote the probability assigned to the predicted verification decision and let $\gamma \in [0, 1]$ represent the confidence threshold. 
Our empirical analysis shows that high-confidence predictions ($p_D \ge \gamma$) are overwhelmingly correct, while low-confidence cases correspond to genuinely ambiguous or out-of-distribution instances.

We evaluate two model families on the MATH500 benchmark~\cite{hendrycks2021measuringmathematicalproblemsolving}: DeepSeek-R1-Distill (1.5B draft, 32B target) and Qwen3 (1.7B draft, 32B target), using target model decisions as ground truth to compare draft verification accuracy (Overall) versus high-confidence instances only (Hi-Conf, $p_D \ge 0.9$). As shown in Figure~\ref{fig:motivation}(a), filtering by confidence substantially improves accuracy: DeepSeek-1.5B improves from 56\% to 81\% (+25\%), while Qwen3-1.7B improves from 71\% to 87\% (+16\%). Importantly, Figure~\ref{fig:motivation}(b) shows that high-confidence predictions cover 61\% and 85\% of all verification instances, respectively. This indicates that the majority of steps can be verified reliably by the draft model alone, while only a small fraction requires escalation.
These findings motivate confidence-gated cascaded verification: accept draft decisions when confident, and defer to the target model otherwise.

\definecolor{comment_color_2}{RGB}{64,128,128}
\newcommand{\LineComment}[1]{\vspace*{0.1em}\fontsize{10}{11}\selectfont\textcolor{comment_color_2}{\textit{\# #1}}}
\begin{algorithm}[t]
\caption{Cascaded Speculative Verification}
\label{alg:cascadespec}
\begin{algorithmic}[1]
\fontsize{10}{11}\selectfont
\REQUIRE Input $x$, draft model $M_D$, target model $M_T$, draft steps $k$, confidence threshold $\gamma$
\ENSURE Reasoning trace $S$
\STATE $S \leftarrow \emptyset$, $\texttt{context} \leftarrow x$

\WHILE{\textbf{not} \texttt{terminated}}
\STATE \LineComment{Stage 1: Step-Level Drafting}
\STATE $\{\hat{s}_1, \ldots, \hat{s}_k\} \leftarrow M_D.\texttt{generate}(\texttt{context}, k)$

\STATE \LineComment{Stage 2: Dual-Tier Cascaded Verification}
\FOR{$j = 1$ \TO $k$ \textbf{in parallel}}
    \STATE $s_j \leftarrow M_T.\texttt{generate}(\texttt{context} \oplus \hat{s}_j)$
\ENDFOR

\STATE $m \leftarrow 0$
\FOR{$j = 1$ \TO $k$}
    \STATE $r_D, p_D \leftarrow M_D.\texttt{verify}(\hat{s}_j, s_j)$
    \STATE $V \leftarrow r_D$ \textbf{if} $p_D \geq \gamma$ \textbf{else} $M_T.\texttt{verify}(\hat{s}_j, s_j)$
    \STATE $m \leftarrow m + 1$ \textbf{if} $V = \texttt{accept}$ \textbf{else} \textbf{break}
\ENDFOR

\STATE \LineComment{Stage 3: Context Management and Fallback}
\IF{$m > 0$}
    \STATE $S \leftarrow S \cup \{\hat{s}_1, \ldots, \hat{s}_{m}\}$
    \STATE $\texttt{context} \leftarrow \texttt{context} \oplus \hat{s}_1 \oplus \cdots \oplus \hat{s}_{m}$
\ELSE
    \STATE $s \leftarrow M_T.\texttt{generate}(\texttt{context})$
    \STATE $S \leftarrow S \cup \{s\}$, $\texttt{context} \leftarrow \texttt{context} \oplus s$
\ENDIF

\ENDWHILE
\RETURN $S$
\end{algorithmic}
\end{algorithm}

\subsection{The \sysname Framework}
\label{sec:framework}

We now formally describe \sysname. Algorithm~\ref{alg:cascadespec} provides the full procedure.

\noindent
\textbf{Notation.}
Let $M_D$ denote a lightweight draft model and $M_T$ a high-capacity target model. Given an input $x$, reasoning proceeds as a sequence of steps $S = \{s_1, \dots, s_n\}$, where each step is a contiguous token span terminated by ``\textbackslash n\textbackslash n''. At step $i$, the draft model proposes candidate steps $\hat{s}_i$, which are verified against the target model's continuation.

\noindent
\textbf{Stage 1: Step-Level Drafting.}
At each iteration, the draft model autoregressively generates $k$ candidate reasoning steps conditioned on the current context (Algorithm~\ref{alg:cascadespec}, line 4):
\begin{equation}
    \setlength{\abovedisplayskip}{3pt}
    \setlength{\belowdisplayskip}{3pt}
\hat{s}_i, \ldots, \hat{s}_{i+k-1} \sim P_{M_D}(\cdot \mid x, s_1, \ldots, s_{i-1}).
\end{equation}
Each draft step is generated until a delimiter is produced, which explicitly defines the step boundary.

\noindent
\textbf{Stage 2: Dual-Tier Cascaded Verification.}
For each draft step $\hat{s}_j$, the target model generates a corresponding step $s_j$ in parallel (Algorithm~\ref{alg:cascadespec}, lines 6--8).
Verification is then performed using a two-tier cascade.

\noindent
\emph{(1) Draft Verification.}
The draft model $M_D$ acts as a binary verifier, taking as input the context, the draft step $\hat{s}_j$, and the target step $s_j$.
It outputs a decision (line 11)
\begin{equation}
    \setlength{\abovedisplayskip}{3pt}
    \setlength{\belowdisplayskip}{3pt}
r_D \in \{\texttt{accept}, \texttt{reject}\},
\end{equation}
along with a confidence score
\begin{equation}
    \setlength{\abovedisplayskip}{3pt}
    \setlength{\belowdisplayskip}{3pt}
p_D = p_{M_D}(r_D \mid x, s_{<i}, \hat{s}_j, s_j).
\end{equation}
In practice, $r_D$ is represented by a special verification token (e.g., ``Yes'' or ``No''), and $p_D$ is the corresponding softmax probability.

\noindent
\emph{(2) Cascaded Decision Rule.}
We define the cascaded verification function as
\begin{equation}
    \setlength{\abovedisplayskip}{3pt}
    \setlength{\belowdisplayskip}{3pt}
V_{\text{cascade}}(\hat{s}_j, s_j) =
\begin{cases}
r_D, & \text{if } p_D \ge \gamma, \\
r_T, & \text{otherwise},
\end{cases}
\end{equation}
where $r_T$ is the target model's verification decision and $\gamma \in [0,1]$ is a confidence threshold. This cascading mechanism preserves target-model accuracy while substantially reducing computational cost (\S\ref{sec:experiments}).
High-confidence draft decisions are accepted directly, while low-confidence cases are escalated to the target model.

As shown in Algorithm~\ref{alg:cascadespec} (lines 12--13), verification proceeds sequentially over the drafted steps until a rejection occurs.

\noindent
\textbf{Stage 3: Context Management and Fallback.}
If the first $m$ draft steps are accepted, they are appended to the reasoning context (lines 16--18).
Upon the first rejection, the target model regenerates the next step from the current context (lines 19--21), ensuring correctness is preserved.
This fallback mechanism guarantees that the final reasoning trace matches the target model’s distribution up to verification errors.

\subsection{Extension: Tree-Structured Drafting}
\label{sec:tree}

\sysname naturally extends to tree-structured speculative reasoning. Instead of generating a single linear sequence, the draft model proposes multiple candidate steps at each layer. Cascaded verification is applied independently to all candidates in parallel.

When multiple candidates pass verification, we select the step with the highest confidence score:
\begin{equation}
    \setlength{\abovedisplayskip}{3pt}
    \setlength{\belowdisplayskip}{3pt}
\hat{s}^* = \arg\max_{\hat{s} \in \mathcal{C}} p(\hat{s}),
\end{equation}
where $p(\hat{s})$ denotes the verification confidence (from $M_D$ or $M_T$ depending on cascading). Beyond verification, this confidence score also serves as a principled criterion for path selection, favoring more reliable reasoning branches.

\subsection{Efficiency Analysis}
\label{sec:analysis}

We analyze the efficiency gains of \sysname. Let $c_{vD}$ and $c_{vT}$ denote the cost of draft and target verification ($c_{vD} \ll c_{vT}$), and let $\alpha \in [0,1]$ be the cascade rate. The expected verification cost per step is $\mathcal{O}(c_{vD} + \alpha \cdot c_{vT})$, approaching draft-only cost when $\alpha$ is small. The threshold $\gamma$ controls this trade-off: higher $\gamma$ improves reliability but increases $\alpha$.

Beyond per-step savings, accurate verification sustains high acceptance rates. Erroneous acceptance of low-quality steps corrupts the reasoning trajectory; while the target model may recover through reflection, the draft model typically cannot—causing repeated rejections as subsequent drafts diverge from the corrupted context, ultimately degrading speedup. By maintaining verification accuracy, \sysname preserves coherent trajectories and high acceptance rates throughout.

Importantly, verification-level cascading is orthogonal to token-level speculative decoding, enabling multiplicative acceleration when combined.
\section{Experiments}
\label{sec:experiments}
\begin{table*}[htbp]
\centering
\definecolor{MyHighlightColor}{RGB}{181,126,220}
\newcommand{\highlightrow}{\rowcolor{MyHighlightColor!15}}

\caption{Main Results: Performance comparison. We report \textbf{Pass@1 Accuracy (\%)} and \textbf{End-to-End Speedup ($\times$)}. The target model baseline represents the upper bound for accuracy and the baseline ($1.00\times$) for speed.}
\label{tab:performance_filled_specreason}
\resizebox{\textwidth}{!}{%
\begin{tabular}{lcccccc}
\toprule
\multirow{2}{*}{\textbf{Method}} & \multicolumn{6}{c}{\textbf{Dataset (Acc. $\uparrow$ / Speedup $\uparrow$)}} \\
\cmidrule(lr){2-7}
& \textbf{AIME24} & \textbf{AMC23} & \textbf{MATH500} & \textbf{GPQA} & \textbf{HumanEval} & \textbf{Mean} \\
\midrule

\multicolumn{7}{c}{\textbf{Draft: Deepseek-R1-Distill 1.5B / Target: Deepseek-R1-Distill 32B}} \\
\midrule

Target Model
& $68.54~/~1.00\times$ & $94.22~/~1.00\times$ & $92.43~/~1.00\times$ & $61.71~/~1.00\times$ & $93.12~/~1.00\times$ & $81.99~/~1.00\times$ \\

\cmidrule[0.3pt](lr){1-7}
SpecReason ($s=7$)
& $38.33~/~1.54\times$ & $80.78~/~2.06\times$ & $78.23~/~1.89\times$ & $51.09~/~1.41\times$ & $75.61~/~1.72\times$ & $64.81~/~1.72\times$ \\

SpecReason ($s=9$)
& $60.21~/~0.82\times$ & $88.91~/~1.06\times$ & $84.61~/~1.13\times$ & $58.59~/~0.76\times$ & $90.17~/~1.05\times$ & $76.50~/~0.96\times$ \\

Embedding ($\alpha=0.8$)
& $61.67~/~1.41\times$ & $92.97~/~1.52\times$ & $90.88~/~1.50\times$ & $56.63~/~1.23\times$ & $88.99~/~1.31\times$ & $78.23~/~1.39\times$ \\

Embedding ($\alpha=0.9$)
& $68.54~/~1.27\times$ & $93.75~/~1.28\times$ & $92.13~/~1.27\times$ & $61.02~/~1.09\times$ & $89.63~/~1.27\times$ & $81.01~/~1.24\times$ \\

LookaheadReasoning
& $67.29~/~1.38\times$ & $94.06~/~1.51\times$ & $92.41~/~1.48\times$ & $60.84~/~1.21\times$ & $91.63~/~1.36\times$ & $81.25~/~1.39\times$ \\

\highlightrow
\textbf{\sysname}
& $\mathbf{68.75~/~1.73\times}$ & $\mathbf{94.53~/~1.76\times}$ & $\mathbf{92.35~/~1.83\times}$ & $\mathbf{60.57~/~1.54\times}$ & $\mathbf{91.33~/~1.53\times}$ & $\mathbf{81.51~/~1.68\times}$ \\

\cmidrule[0.3pt](lr){1-7}

SD (Lossless)
& $\equiv~/~1.48\times$ & $\equiv~/~1.46\times$ & $\equiv~/~1.43\times$ & $\equiv~/~1.52\times$ & $\equiv~/~1.37\times$ & $\equiv~/~1.45\times$ \\

\highlightrow
\textbf{\sysname + SD}
& $\mathbf{\equiv^*~/~2.12\times}$ & $\mathbf{\equiv^*~/~2.03\times}$ & $\mathbf{\equiv^*~/~2.24\times}$ & $\mathbf{\equiv^*~/~2.01\times}$ & $\mathbf{\equiv^*~/~1.81\times}$ & $\mathbf{\equiv^*~/~2.04\times}$ \\

\midrule
\midrule 

\multicolumn{7}{c}{\textbf{Draft: Qwen3 1.7B / Target: Qwen3 32B}} \\
\midrule

Target Model
& $80.21~/~1.00\times$ & $97.34~/~1.00\times$ & $92.54~/~1.00\times$ & $67.83~/~1.00\times$ & $93.68~/~1.00\times$ & $86.32~/~1.00\times$ \\

\cmidrule[0.3pt](lr){1-7}

SpecReason ($s=7$)
& $40.21~/~1.29\times$ & $83.13~/~1.53\times$ & $82.40~/~1.54\times$ & $54.24~/~1.06\times$ & $79.87~/~1.37\times$ & $67.97~/~1.36\times$ \\

SpecReason ($s=9$)
& $69.38~/~1.08\times$ & $91.41~/~1.11\times$ & $87.29~/~1.33\times$ & $63.18~/~0.77\times$ & $91.15~/~1.15\times$ & $80.48~/~1.09\times$ \\

Embedding ($\alpha=0.8$)
& $70.42~/~1.23\times$ & $95.16~/~1.24\times$ & $91.67~/~1.30\times$ & $66.23~/~1.13\times$ & $91.06~/~1.15\times$ & $82.91~/~1.21\times$ \\

Embedding ($\alpha=0.9$)
& $74.17~/~1.10\times$ & $95.83~/~1.14\times$ & $91.72~/~1.17\times$ & $66.41~/~1.03\times$ & $92.07~/~1.06\times$ & $84.04~/~1.10\times$ \\

LookaheadReasoning
& $79.58~/~1.19\times$ & $96.41~/~1.24\times$ & $92.40~/~1.28\times$ & $67.17~/~1.10\times$ & $92.73~/~1.17\times$ & $85.66~/~1.20\times$ \\

\highlightrow
\textbf{\sysname}
& $\mathbf{79.79~/~1.27\times}$ & $\mathbf{96.41~/~1.35\times}$ & $\mathbf{92.82~/~1.42\times}$ & $\mathbf{67.53~/~1.18\times}$ & $\mathbf{94.35~/~1.27\times}$ & $\mathbf{86.18~/~1.30\times}$ \\

\cmidrule[0.3pt](lr){1-7}

SD (Lossless)
& $\equiv~/~1.39\times$ & $\equiv~/~1.36\times$ & $\equiv~/~1.34\times$ & $\equiv~/~1.39\times$ & $\equiv~/~1.39\times$ & $\equiv~/~1.37\times$ \\

\highlightrow
\textbf{\sysname + SD}
& $\mathbf{\equiv^*~/~1.62\times}$ & $\mathbf{\equiv^*~/~1.73\times}$ & $\mathbf{\equiv^*~/~1.79\times}$ & $\mathbf{\equiv^*~/~1.54\times}$ & $\mathbf{\equiv^*~/~1.58\times}$ & $\mathbf{\equiv^*~/~1.65\times}$ \\

\bottomrule
\end{tabular}
}
\caption*{\footnotesize \textit{Note}: `$\equiv$' denotes lossless accuracy matching the target model. `$\equiv^*$' denotes accuracy matching the base method.}
\vspace{-5mm}
\end{table*}

\subsection{Experimental Setup}

\noindent
\textbf{Models.}
We evaluate \sysname on two widely used open-source model families with long-CoT reasoning capabilities: DeepSeek-R1-Distill\cite{deepseekai2025deepseekr1incentivizingreasoningcapability} and Qwen3\cite{yang2025qwen3technicalreport}. For both families, we fix the target model to the 32B variant and use a substantially smaller draft model (1.5B for DeepSeek-R1-Distill and 1.7B for Qwen3). This setting reflects a realistic deployment scenario where a lightweight model assists a high-capacity reasoning model, while keeping the target model identical across all methods.

\noindent
\textbf{Datasets.}
We consider five benchmarks spanning mathematical reasoning, scientific question answering, and code generation.
Specifically, we evaluate on AIME24~\cite{aime24}, AMC23~\cite{amc23}, and MATH500~\cite{hendrycks2021measuringmathematicalproblemsolving} for mathematical reasoning; GPQA Diamond~\cite{rein2023gpqagraduatelevelgoogleproofqa} for graduate-level scientific QA; and HumanEval~\cite{chen2021codex} for code generation.
These datasets require long intermediate reasoning chains and cover diverse paradigms, enabling assessment of domain generalization.

\noindent
\textbf{Generation Parameters.}
Unless otherwise specified, all methods use the official recommended configurations of their respective model families. We fix the maximum token budget to 32K for DeepSeek-R1-Distill and 37K for Qwen3 to ensure complete reasoning traces. For step-level speculative methods, We set the speculation depth to 5 steps per iteration for \sysname, while baselines follow their original settings. All generation parameters are kept identical across methods to ensure fair comparison.

\noindent
\textbf{Baselines.}
We compare against vanilla target-model inference as the primary accuracy and speedup baseline. In addition, we evaluate three representative step-level speculative reasoning methods: 
(i) \textbf{SpecReason}, which performs scoring-based verification (evaluated at score thresholds $s \in \{7, 9\}$);
(ii) \textbf{LookaheadReasoning}, which relies on an external LLM-as-a-Judge for semantic equivalence verification; and (iii) \textbf{Embedding Alignment}, which verifies steps via embedding-based semantic similarity (evaluated at thresholds $\alpha \in \{0.8, 0.9\}$). To demonstrate orthogonality with token-level speculative decoding, we also report results for \sysname+ speculative decoding (SD), which combines our method with n-gram-based token-level speculation (prompt lookup decoding~\citealp{saxena2023prompt}) in the main experiments, 
with additional methods such as EAGLE-3~\cite{li2025eagle3scalinginferenceacceleration} discussed in Appendix~\ref{sec:appendix_eagle}. 
All baselines are evaluated under the same token budgets and hardware constraints.

\noindent
\textbf{Evaluation.}
All experiments are conducted on NVIDIA H200 GPUs. The target model is deployed using tensor parallelism across two GPUs, while the draft model runs on a single GPU. Importantly, \sysname does not require additional hardware resources beyond those used by the target and draft models. Our implementation builds on vLLM~0.10.1~\cite{vllm}, and all methods are evaluated within the same serving framework. Following prior work~\cite{pan2025specreason,fu2025scaling}, 
we report pass@1 accuracy (averaged over 16 samples per problem at temperature 0.6),
along with relative inference speedup normalized to the target-model baseline.

\subsection{Main Results}
\paragraph{Pareto Frontier of Accuracy and Speed.}
The core challenge in speculative reasoning is not a smooth accuracy--latency trade-off, but the risk of sharp and irreversible accuracy collapse under aggressive verification shortcuts.
Table~\ref{tab:performance_filled_specreason} summarizes the resulting accuracy and speed characteristics.
\begin{itemize}
\item \textbf{SpecReason:} Aggressive thresholds yield substantial speedups (e.g., $1.72\times$ with $s=7$), but cause severe accuracy degradation on complex tasks (e.g., from $68.46\%$ to $38.33\%$ on AIME24). This failure stems from utility-based scoring that favors fluent yet semantically incorrect reasoning paths, which irreversibly misguides subsequent generation.

\item \textbf{Embedding:} Despite avoiding additional inference engines, embedding-based verification struggles to capture semantic equivalence beyond lexical similarity. Consequently, logically consistent but paraphrased steps are frequently misclassified, leading to a reduced accuracy of $78.23\%$ at $\alpha=0.8$ despite a modest $1.39\times$ speedup.

\item \textbf{LookaheadReasoning:} This method preserves accuracy most effectively among baselines, closely matching the target model. However, its dependence on an external LLM-as-a-Judge incurs substantial computational redundancy, limiting its practical efficiency gains.

\item \textbf{\sysname (Ours):} 
\sysname achieves a superior Pareto frontier, matching the target model’s accuracy distribution while delivering significant acceleration. It attains average speedups of $1.68\times$ and $1.30\times$ on the DeepSeek and Qwen families, respectively. This indicates that our confidence-gated cascade effectively identifies "easy" steps that the draft model can handle reliably, while correctly routing "complex" reasoning steps to the target model for rigorous verification.
\end{itemize}

\paragraph{Orthogonality with Token-Level Speculation.}
Table~\ref{tab:performance_filled_specreason} demonstrates that combining \sysname with speculative decoding (SD) yields multiplicative speedup gains. On DeepSeek-R1-Distill, \sysname+ SD achieves 2.04$\times$ mean speedup while preserving base method accuracy, compared to 1.45$\times$ for standalone SD. On Qwen3, the combination achieves 1.65$\times$ speedup versus 1.37$\times$ for SD alone. These results validate that \sysname is fully orthogonal to token-level speculation, enabling the stacked acceleration as discussed in Section~\ref{sec:analysis}.

\paragraph{Generalization and Robustness.}
\sysname exhibits robust and consistent improvements across both model families and a wide range of task domains, including mathematical reasoning, scientific question answering, and code generation. While absolute speedups vary across datasets (1.53$\times$--1.83$\times$ on DeepSeek, 1.18$\times$--1.42$\times$ on Qwen3), this variation is expected and reflects differences in reasoning complexity and step acceptance rates rather than method fragility. Differences across model families primarily stem from draft-model capability and alignment with the target model. Crucially, despite these variations, the relative ranking of methods remains consistent across all settings, with \sysname achieving the best accuracy--speed trade-off in every configuration.

\subsection{Ablation Studies}

\begin{table*}[htbp]
\centering
\definecolor{MyHighlightColor}{RGB}{181,126,220}
\newcommand{\highlightrow}{\rowcolor{MyHighlightColor!15}}
\caption{Ablation study on verification strategies. We compare static strategies (Draft/Target Only) with \sysname under different confidence thresholds $\gamma$. Cell format: \textbf{Accuracy / Speedup}.}
\label{tab:performance_target_verl_final_update_v4}
\resizebox{\textwidth}{!}{%
\begin{tabular}{lcccccc}
\toprule
\multirow{2}{*}{\textbf{Method}} & \multicolumn{6}{c}{\textbf{Dataset (Acc. $\uparrow$ / Speedup $\uparrow$)}} \\
\cmidrule(lr){2-7}
& \textbf{AIME24} & \textbf{AMC23} & \textbf{MATH500} & \textbf{GPQA} & \textbf{HumanEval} & \textbf{Mean} \\
\midrule

Target Model
& $68.46~/~1.00\times$ & $94.22~/~1.00\times$ & $92.43~/~1.00\times$ & $61.71~/~1.00\times$ & $93.12~/~1.00\times$ & $82.03~/~1.00\times$ \\

\cmidrule[0.3pt](lr){1-7}

Draft Verification
& $35.89~/~2.34\times$ & $73.91~/~2.18\times$ & $81.92~/~2.19\times$ & $43.03~/~2.15\times$ & $69.68~/~2.00\times$ & $60.89~/~2.17\times$ \\

\sysname ($\gamma=0.8$)
& $49.38~/~1.94\times$ & $82.97~/~1.86\times$ & $90.60~/~1.96\times$ & $56.05~/~1.68\times$ & $84.14~/~1.70\times$ & $72.63~/~1.83\times$ \\

\highlightrow
\textbf{\sysname ($\gamma=0.9$)}
& $\mathbf{68.75~/~1.73\times}$ & $\mathbf{94.53~/~1.76\times}$ & $\mathbf{92.35~/~1.83\times}$ & $\mathbf{60.57~/~1.54\times}$ & $\mathbf{91.33~/~1.53\times}$ & $\mathbf{81.51~/~1.68\times}$ \\

\sysname ($\gamma=0.95$)
& $68.46~/~1.54\times$ & $94.69~/~1.72\times$ & $92.35~/~1.72\times$ & $61.36~/~1.36\times$ & $91.42~/~1.46\times$ & $81.66~/~1.56\times$ \\

Target Verification
& $68.75~/~1.42\times$ & $94.84~/~1.48\times$ & $92.43~/~1.59\times$ & $61.47~/~1.27\times$ & $91.39~/~1.35\times$ & $81.78~/~1.42\times$ \\

\bottomrule
\end{tabular}%
}
\end{table*}
\begin{table*}[htbp]
\centering
\definecolor{MyHighlightColor}{RGB}{181,126,220}
\newcommand{\highlightrow}{\rowcolor{MyHighlightColor!15}}

\caption{Ablation study on tree width $W$. Cell format: \textbf{Accuracy / Speedup}.}
\label{tab:performance_filled_w_variants}
\resizebox{\textwidth}{!}{
\begin{tabular}{lcccccc}
\toprule
\multirow{2}{*}{\textbf{Method}} & \multicolumn{6}{c}{\textbf{Dataset (Acc. $\uparrow$ / Speedup $\uparrow$)}} \\
\cmidrule(lr){2-7}
& \textbf{AIME24} & \textbf{AMC23} & \textbf{MATH500} & \textbf{GPQA} & \textbf{HumanEval} & \textbf{Mean} \\
\midrule

Target Model
& $68.54~/~1.00\times$ & $94.22~/~1.00\times$ & $92.43~/~1.00\times$ & $61.71~/~1.00\times$ & $93.12~/~1.00\times$ & $82.00~/~1.00\times$ \\

\cmidrule[0.3pt](lr){1-7}
\highlightrow
\textbf{\sysname ($W=1$)}
& $\mathbf{68.75~/~1.73\times}$ & $\mathbf{94.53~/~1.76\times}$ & $\mathbf{92.35~/~1.83\times}$ & $\mathbf{60.57~/~1.54\times}$ & $\mathbf{91.33~/~1.53\times}$ & $\mathbf{81.51~/~1.68\times}$ \\

\sysname ($W=2$)
& $68.54~/~1.39\times$ & $94.69~/~1.56\times$ & $92.35~/~1.63\times$ & $60.81~/~1.28\times$ & $91.99~/~1.38\times$ & $81.68~/~1.45\times$ \\

\sysname ($W=4$)
& $68.96~/~0.70\times$ & $95.16~/~0.61\times$ & $92.84~/~0.77\times$ & $60.92~/~0.83\times$ & $92.27~/~0.65\times$ & $82.03~/~0.71\times$ \\

\bottomrule
\end{tabular}%
}
\vspace{-1mm}
\end{table*}
\paragraph{Effect of Verification Strategy and Threshold.}
We investigate the impact of different verification strategies and confidence thresholds on the accuracy-speedup trade-off. Table~\ref{tab:performance_target_verl_final_update_v4} presents results comparing: (1) \textbf{Draft Verification}, which exclusively uses the draft model for all verification decisions; (2) \textbf{Target Verification}, which always escalates to the target model; and (3) \sysname with varying confidence thresholds $\gamma \in \{0.8, 0.9, 0.95\}$.

The baselines expose the limitations of relying on a single model. Draft-only verification achieves the highest raw speedup ($2.17\times$) but suffers from catastrophic accuracy degradation, particularly on complex reasoning tasks like AIME24 (dropping from $68.46\%$ to $35.89\%$). This result highlights that the small model lacks sufficient reasoning capability to reliably adjudicate complex logical steps, often misclassifying invalid paths as correct. In contrast, Target-only verification ensures rigorous correctness (Mean Acc. $81.78\%$) but limits the efficiency gain ($1.42\times$), as it incurs high computational costs by activating the large model for every verification instance regardless of its difficulty.

The confidence threshold $\gamma$ provides fine-grained control over the speculation aggressiveness, allowing \sysname to navigate the space between these two extremes. Lower threshold ($\gamma=0.8$) adopts an aggressive policy, boosting speedup to $1.83\times$ but compromising reliability (Mean Acc. drops to $72.63\%$) by allowing the draft model to accept plausible but incorrect steps. Higher threshold ($\gamma=0.95$) forces a conservative behavior, recovering accuracy to near-perfect levels ($81.66\%$) but reducing speedup to $1.56\times$ due to frequent cascading. Our default setting of $\gamma=0.9$ achieves the optimal balance, maintaining $99.7\%$ of the target model's accuracy while delivering a robust $1.68\times$ mean speedup, effectively filtering "easy" steps from "hard" ones.

\paragraph{Effect of Tree-Structured Drafting.} We investigate the impact of tree width on the accuracy-speedup trade-off by evaluating branching factors $W \in \{1, 2, 4\}$. \sysname extends to tree-structured speculation, generating multiple candidate branches at each step. While a larger $W$ theoretically captures superior reasoning paths via confidence-based selection, the number of candidates grows exponentially with depth, imposing substantial overhead for parallel verification.

Table~\ref{tab:performance_filled_w_variants} presents the results of this comparison. Increasing the tree width to $W=4$ indeed yields a marginal accuracy improvement (Mean Acc. rises from $81.51\%$ to $82.03\%$), matching or even marginally exceeding the target model on some tasks. This confirms that a wider search space helps capture superior reasoning trajectories. However, this gain comes at a prohibitive computational cost: the overhead of verifying multiple long steps in parallel overwhelms the speed gains from speculation, resulting in a net slowdown ($0.71\times$ speedup). Even a modest width of $W=2$ significantly dampens the acceleration, reducing it to $1.45\times$ compared to the linear baseline's $1.68\times$.

These results suggest that for current model configurations, linear drafting ($W=1$) provides the optimal trade-off. Tree-structured drafting may become more beneficial when draft model quality is lower or when reasoning tasks require extensive exploration, as the increased candidate diversity can compensate for individual draft quality limitations.

\section{Conclusion}
We revisited step-level speculative reasoning from the perspective of verification and showed that treating all steps as uniformly hard leads to unnecessary computation. Observing that verification difficulty is heterogeneous and draft-model confidence can guide selective escalation, we proposed \sysname, a confidence-gated framework that escalates uncertain steps to the target model. Experiments across workloads show that \sysname matches target-model accuracy while achieving substantial inference speedups, remaining orthogonal to token-level speculative decoding.
\newpage
\section{Limitations}

\sysname relies on draft-model confidence to route step-level verification. While informative in our experiments, this signal may not generalize across all model families, architectures, or task distributions. Step-level verification approximates semantic equivalence via discriminative judgment rather than exact logical validation, which can be brittle for highly nuanced or implicitly structured reasoning steps. Efficiency gains also depend on alignment between the draft and target models; under substantial domain shifts or consistently low draft-model confidence, the system may frequently fall back to target-model inference, limiting speedups. Future work includes systematically characterizing confidence calibration across models, improving verification for nuanced reasoning, and extending \sysname to diverse domains.

\bibliography{custom}
\clearpage
\appendix

\begin{table*}[t]
\centering
\definecolor{MyHighlightColor}{RGB}{181,126,220}
\newcommand{\highlightrow}{\rowcolor{MyHighlightColor!15}}

\caption{Compatibility with token-level speculative decoding methods. We report \textbf{Pass@1 Accuracy (\%)} and \textbf{End-to-End Speedup ($\times$)}. The target model baseline represents the upper bound for accuracy and the baseline ($1.00\times$) for speed.}
\label{tab:eagle}
\resizebox{\textwidth}{!}{%
\begin{tabular}{lcccccc}
\toprule
\multirow{2}{*}{\textbf{Method}} & \multicolumn{6}{c}{\textbf{Dataset (Acc. $\uparrow$ / Speedup $\uparrow$)}} \\
\cmidrule(lr){2-7}
& \textbf{AIME24} & \textbf{AMC23} & \textbf{MATH500} & \textbf{GPQA} & \textbf{HumanEval} & \textbf{Mean} \\
\midrule
\multicolumn{7}{c}{\textbf{Draft: Qwen3 1.7B / Target: Qwen3 32B}} \\
\midrule

Target Model
& $80.21~/~1.00\times$ & $97.34~/~1.00\times$ & $92.54~/~1.00\times$ & $67.83~/~1.00\times$ & $93.68~/~1.00\times$ & $86.32~/~1.00\times$ \\

\highlightrow
\textbf{\sysname}
& $\mathbf{79.79~/~1.27\times}$ & $\mathbf{96.41~/~1.35\times}$ & $\mathbf{92.82~/~1.42\times}$ & $\mathbf{67.53~/~1.18\times}$ & $\mathbf{94.35~/~1.27\times}$ & $\mathbf{86.18~/~1.30\times}$ \\
\cmidrule[0.3pt](lr){1-7}

EAGLE-3
& $\equiv~/~0.96\times$ & $\equiv~/~0.96\times$ & $\equiv~/~0.98\times$ & $\equiv~/~0.89\times$ & $\equiv~/~0.92\times$ & $\equiv~/~0.94\times$ \\

\highlightrow
\textbf{\sysname + EAGLE-3}
& $\mathbf{\equiv^*~/~1.07\times}$ & $\mathbf{\equiv^*~/~1.04\times}$ & $\mathbf{\equiv^*~/~1.06\times}$ & $\mathbf{\equiv^*~/~1.02\times}$ & $\mathbf{\equiv^*~/~1.09\times}$ & $\mathbf{\equiv^*~/~1.06\times}$ \\
\cmidrule[0.3pt](lr){1-7}

PLD
& $\equiv~/~1.39\times$ & $\equiv~/~1.36\times$ & $\equiv~/~1.34\times$ & $\equiv~/~1.39\times$ & $\equiv~/~1.39\times$ & $\equiv~/~1.37\times$ \\

\highlightrow
\textbf{\sysname + PLD}
& $\mathbf{\equiv^*~/~1.62\times}$ & $\mathbf{\equiv^*~/~1.73\times}$ & $\mathbf{\equiv^*~/~1.79\times}$ & $\mathbf{\equiv^*~/~1.54\times}$ & $\mathbf{\equiv^*~/~1.58\times}$ & $\mathbf{\equiv^*~/~1.65\times}$ \\

\bottomrule
\end{tabular}%
}
\caption*{\footnotesize \textit{Note}: `$\equiv$' denotes lossless accuracy matching the target model. `$\equiv^*$' denotes accuracy matching the base method.}
\end{table*}

\section{Compatibility with Token-Level Speculative Decoding}
\label{sec:appendix_eagle}
We provide a detailed analysis of \sysname's compatibility with different token-level speculative decoding methods, comparing Prompt Lookup Decoding and EAGLE-3 to demonstrate how the choice of token-level method impacts overall acceleration.

\subsection{Setup}

We evaluate two representative token-level speculative decoding approaches: (1) \textbf{Prompt Lookup Decoding (PLD)}, a draft-model-free method that maintains a cache of historical tokens and performs n-gram string matching to propose speculative drafts. At each decoding step, it identifies historical n-grams starting with the current sequence's suffix as candidates, which are then verified by the target model in a single forward pass. (2) \textbf{EAGLE-3}, a draft-model-based method that trains a lightweight draft head by leveraging multi-layer features from the target model. This alignment training enables close approximation of the target model's token distribution, achieving state-of-the-art token-level speculative decoding performance on multi-turn dialogue and short-form generation benchmarks.

We conduct experiments using Qwen3 1.7B / 32B as the draft / target model pair, comparing PLD and EAGLE-3 both as standalone methods and in combination with \sysname. All methods are evaluated on the same five benchmarks under identical hardware and generation configurations as described in the main experiments. Results are presented in Table~\ref{tab:eagle}.

\subsection{Results and Analysis}

\paragraph{PLD.}
PLD demonstrates strong standalone performance, achieving $1.37\times$ mean speedup while preserving lossless accuracy across all benchmarks. This effectiveness stems from its cache-based design that dynamically adapts to the generation context, effectively exploiting the repetitive patterns inherent in long chain-of-thought reasoning. Mathematical derivations and structured reasoning traces naturally contain recurring phrases, notation, and expression patterns that PLD's n-gram matching mechanism can efficiently capture and reuse.

\paragraph{EAGLE-3.}
Contrary to its strong results on conversational benchmarks, EAGLE-3 produces a net slowdown ($0.94\times$) on our reasoning tasks. We attribute this to a distributional mismatch between its training data and the nature of complex reasoning. EAGLE-3 is primarily optimized for short-form, predictable conversational tasks. In contrast, long-form mathematical reasoning involves multi-step logical derivations and specialized notation, with sequences frequently exceeding 16K tokens. This distributional shift causes the draft head to be poorly calibrated, leading to high rejection rates. Ultimately, the computational overhead of rejecting these drafts outweighs the gains from valid speculations.

\paragraph{Orthogonality.}
\sysname + PLD achieves multiplicative speedup gains ($1.65\times$), confirming that n-gram-based methods complement step-level speculation effectively by adapting to the repetitive structure of reasoning traces. In contrast, \sysname + EAGLE-3 ($1.06\times$) barely overcomes EAGLE-3's standalone slowdown, as EAGLE-3's inefficiency propagates through the entire pipeline: both target model generation steps and step-level verification are subject to high rejection rates, creating cascading degradation.

These results highlight that the choice of token-level technique should account for task characteristics. Draft-model-free methods like PLD demonstrate greater robustness for complex reasoning, while draft-model-based methods require distributional alignment with the target domain to avoid negating potential speedup gains.

\end{document}